\RequirePackage{fix-cm}
\documentclass[final]{svjour3}       
\usepackage[compact]{titlesec}
\smartqed  
\usepackage[margin=3cm]{geometry}
\usepackage[linesnumbered]{algorithm2e}
\usepackage{graphicx}
\graphicspath{ {./Figures/} }
\usepackage{caption, copyrightbox}
\usepackage{subcaption}
\usepackage{booktabs} 
\usepackage{xparse}   

\usepackage{enumitem}

\usepackage{geometry}
\usepackage[fontsize=12pt]{fontsize}

\usepackage{algpseudocode}
\def\makeheadbox{\relax}
\usepackage{gensymb}
\usepackage{amsmath}
\usepackage{amssymb}
\usepackage{hyperref}
\newcommand\tab[1][1cm]{\hspace*{#1}}
\newcommand{\D}{\mathbb{D}}
\newcommand{\V}{\mathbb{V}}
\newcommand{\F}{\mathbb{F}}

\newcommand{\E}{\mathbb{E}}
\renewcommand{\S}{\mathbb{S}}
\newcommand{\Z}{\mathbb{Z}}
\newcommand{\R}{\mathbb{R}}

\newcommand{\collection}[1]{\ensuremath{\left\{#1 \right\}}}
\newcommand{\fof}[1]{\ensuremath{\left(#1\right)}}
\DeclareMathOperator*{\argmax}{argmax}

\usepackage[numbers,sort&compress]{natbib}

\begin{document}


\title{Multi-Resolution Factor Graph Based Stereo Correspondence Algorithm
}
\author{}
\subtitle{Hanieh Shabanian, Madhusudhanan Balasubramanian}

\titlerunning{Multi-Resolution Factor Graph Based Stereo Correspondence Algorithm}        



\institute{H. Shabanian \email{hshbnian@memphis.edu} \and M. Balasubramanian \email{mblsbrmn@memphis.edu} \\ Department of Electrical and Computer Engineering, The University of Memphis} 

\date{}

\maketitle
\vspace{-0.5in}
\begin{abstract} 
A dense depth-map of a scene at an arbitrary view orientation can be estimated from dense view correspondences among multiple lower-dimensional views of the scene.  These low-dimensional view correspondences are dependent on the geometrical relationship among the views and the scene. Determining dense view correspondences is difficult in part due to presence of homogeneous regions in the scene and due to presence of occluded regions and illumination differences among the views. We present a new multi-resolution factor graph-based stereo matching algorithm (MR-FGS) that utilizes both intra- and inter-resolution dependencies among the views as well as among the disparity estimates. The proposed framework allows exchange of information among multiple resolutions of the correspondence problem and is useful for handling larger homogeneous regions in a scene. The MR-FGS algorithm was evaluated qualitatively and quantitatively using stereo pairs in the \textit{Middlebury stereo benchmark dataset} based on commonly used performance measures. When compared to a recently developed factor graph model (FGS), the MR-FGS algorithm provided more accurate disparity estimates without requiring the commonly used post-processing procedure known as the \textit{left-right consistency check}.  The multi-resolution dependency constraint within the factor-graph model significantly improved contrast along depth boundaries in the MR-FGS generated disparity maps. 


\end{abstract}

\section*{Declarations}
    \underline{Funding}: This research was supported in part by an unrestricted start-up fund from the Herff College of Engineering and a graduate assistantship from the Department of Electrical and Computer Engineering, The University of Memphis.\\
    \underline{Conflicts of interest}: None.\\
    \underline{Availability of data and material}: Public benchmark datasets\\
    \underline{Code availability}: Available upon request

\section{Introduction and Related Work}
\label{intro}

    Depth profile of a scene can be estimated from a map of pixel-wise correspondences or co-ordinate disparities among the stereo views of the scene using the inverse depth-disparity relationship.  Estimates of scene stereo disparity are useful for generating dense 3D geometries and architectures of all the elements in the scene.  Such 3D reconstructions and depth estimates are useful for robotic navigation \cite{desouza2002vision}, 3D surface reconstruction \cite{remondino2008turning}, route planning and autonomous navigation \cite{shean2016automated}, remote sensing \cite{shean2016automated}, augmented reality \cite{zenati2007dense}, and for object detection \cite{helmer2010using}.  In general, observations or images acquired using a stereo imaging system are dependent on the camera optics including their focal lengths, distance between the two cameras (baseline distance) and the locations as well as the nature of the illumination sources.  Determining dense stereo correspondence from a stereo image pair is an ill-posed inverse problem due to presence of occlusion, homogeneity and illumination variation among the multiple views of the scene.  
    
    Based on the taxonomy of stereo correspondence algorithms, stereo matching methods can be categorized in two broad groups namely energy-based and window-based algorithms \cite{scharstein2002taxonomy}. The energy-based methods are also called global methods as their cost or objective functions are defined as a function of the entire image extents \cite{yang2008stereo}. In contrary, window-based algorithms utilizes fixed or adaptive windows to define a cost function based on smaller pixel neighborhoods. Although the local-based stereo matching algorithms are suitable for real-time applications, the local windowed-methods have lower estimation accuracy in the presence of occluded regions among the multiple views and when there are scene elements with homogeneous textures (for example, a scene with a homogeneous background). 
    
    Utilizing spatial dependencies of the stereo disparity estimates as well as of the scene characteristics are proven strategies for improving the accuracy of the stereo disparity estimates. In Markov random field (MRF) \cite{li1994markov} based methods, random variables satisfying the Markov property \cite{ethier2009markov} are defined on the pixel lattice to model unknown disparities within each stereo image pair. The joint distribution of the random field is then defined as function of the conditional distribution of each of the random variables. Though MRF-based stereo matching algorithms improve the accuracy of the disparity estimates, they require \textit{maximal} spatial dependencies among pixels in the chosen MRF neighborhood system. Further, because the neighborhood system is uniformly enforced for all the pixel locations, pairwise cliques or $2 \times 2$ cliques are commonly used in these MRF models. Previously, we presented a new factor graph-based probabilistic graphical model (FGS) that addresses these limitations of MRF-based disparity optimization techniques \cite{shabanian2021novel}. More specifically, the FGS method allows spatially variable, larger and computationally optimal neighborhood systems for probabilistic graphical models.  The FGS algorithm provided disparity estimates with higher accuracy when compared to recent non-learning as well as learning-based disparity estimation algorithms. 
    
    A multi-resolution computational framework provides a hierarchical descriptions of any mathematical function as a function of successive lower dimensions of its domain \cite{farin2002handbook}. Thus, the function is represented at multiple dimensions of its domain or at multiple resolutions ranging from the finest to the coarsest resolution of its domain \cite{navarro1996image}. Such multi-resolution representations are useful for reducing the computational complexity of algorithms and for accessing objects or elements in a scene at various scales. Multi-resolution techniques are widely used in various imaging and computational applications such as image segmentation \cite{salem2008multiresolution}, image manipulation \cite{shabanian2017new}, motion analysis \cite{tosic2005multiresolution}, and for stereo depth estimation \cite{zhao2019super}. In disparity estimation, the multi-resolution approach bridges the local window-based and the global energy-based approaches by improving the accuracy and convergence of disparity estimation algorithms. 
    
    Several multi-layer MRF models have been proposed for computer vision problems such as the factorial MRF model \cite{kim2002factorial} extended from the standard MRF model; a multi-layer MRF model for segmenting textured color images \cite{kato2002multicue}; MRF models for change detection in optical remote sensing images \cite{benedek2015multilayer}; and for detecting object motion regions in aerial images \cite{benedek2009detection}. Among the MRF-based multi-layer graphical models, fewer models utilized multi-level factor graphs-based structures. In \cite{shi2007factor}, Shi and You presented a multilevel factor graph-based meta-model for sensor fusion. In their work, the sensor fusion problem was modeled to account for influence of weather conditions on tractable objects in a wireless network. To obtain higher order dependencies among random variables, Zhan and Wu \cite{zhang2019factor} extended the graph-based neural networks \cite{gilmer2017neural} to factor graph-based neural networks. Recently, a general multi-level factor graph framework was developed for modeling variables with multiple levels of dependencies with demonstrated application in image defogging \cite{mutimbu2018factor}. In this framework, unknown variables were estimated using maximum \textit{a posteriori} inference following convergence of a max-product message passing procedure. While benefits of multi-layer factor graphs are evident, to the best of our knowledge, factor graph models have not been previously formulated for solving multi-view correspondence problems.  
    
    In this paper, we present a multi-resolution factor graph (MR-FGS) model that uses spatial dependencies as well as multi-resolution dependencies among random variables in probabilistic graphical models. While the model has broader applications, we specifically present the MR-FGS model for stereo disparity estimation with demonstrated improvement in the disparity estimation accuracy over the FGS model and improved contrast along the depth boundaries. We compared the performance of the proposed multi-resolution probabilistic factor graph model with FGS results by conducting extensive experiments using the Middlebury benchmark stereo datasets \cite{scharstein2003high}, \cite{scharstein2007learning}, \cite{hirschmuller2007evaluation}, \cite{scharstein2014high}.
    
    The rest of the paper is organized as follows. A detailed description of the multi-resolution  probabilistic factor graph-based stereo disparity estimation (MR-FGS) algorithm is presented in Section~\ref{sec:MR-FG model}. The experiments and comparisons are discussed in Section~\ref{Experiments_MRFGS}. Our conclusions for the MR-FGS model are presented in Section~\ref{Conclusions_MRFGS}.

\section{Probabilistic Multi-Resolution Factor Graph Model for Disparity Estimation}
\label{sec:MR-FG model}
Let, each of the images in the stereo pair be of size $M \times N$ pixels at the original resolution and of size $\frac{M}{\alpha} \times \frac{N}{\alpha}$ at a resolution level $\zeta \in \Z_0^{+}$, where $\Z_0^{+}$ represents a collection of all non-negative integers and $\alpha = 2^\zeta$.  Thus, the resolution level $\zeta = 0$ represents the original resolution and the resolution is progressively scaled by a factor of 2 at resolutions levels $\zeta > 0$.  For notational convenience, each pixel coordinate at a level $\zeta$ in the multi-resolution pyramid  is referred using a linear index $i \in \collection{1, \ldots, \frac{M}{\alpha} \frac{N}{\alpha}: \alpha = 2^\zeta; \zeta \subset \Z^{+}_{0}}$.
\subsection{Graph Structure}
\label{sec:MRFG structure}
\begin{figure}
        \includegraphics[width=0.90\textwidth,keepaspectratio]{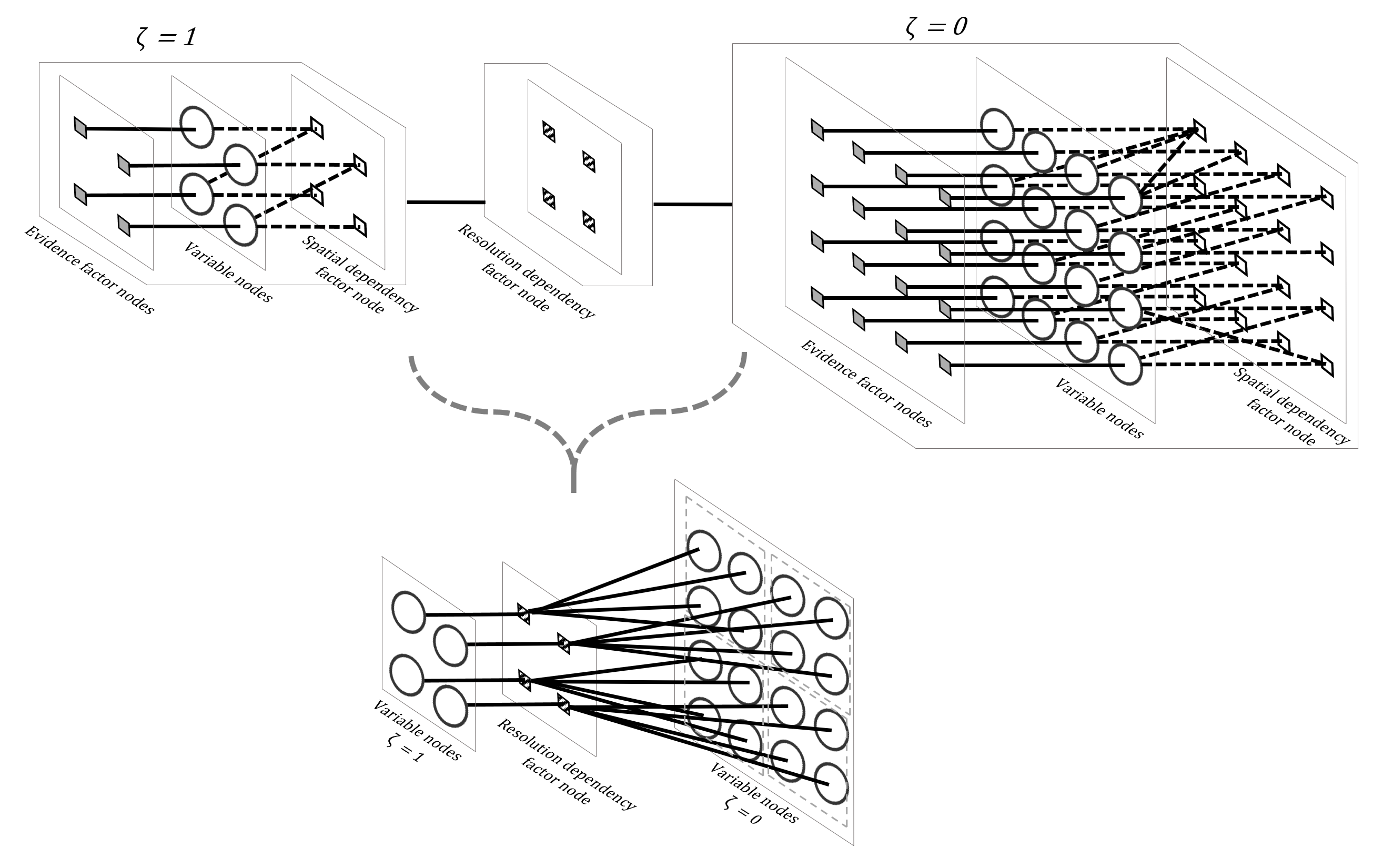}\centering
        \caption{Schematic representation of the proposed multi-resolution factor graph model (with two resolution levels chosen for illustration) designed for optimal estimation of dense stereo disparities. Let, each image in the stereo pair be of of size $M \times N$ pixels.  At the original resolution level $\zeta = 0$ in the multi-resolution factor graph, there are $MN$ number of variable nodes (circular nodes) $\V^{\zeta=0}$, $MN$ number of dependency factor nodes (empty square nodes) $\S^{\zeta=0}$, and $MN$ number of evidence factor nodes (solid square nodes) $\E^{\zeta=0}$.  At the second resolution level $\zeta = 1$ in the multi-resolution factor graph, there are $\frac{M}{2} \frac{N}{2}$ variable nodes (circular nodes) $\V^{\zeta=1}$; $\frac{M}{2} \frac{N}{2}$ number of dependency factor nodes (empty square nodes) $\S^{\zeta=1}$ and $\frac{M}{2} \frac{N}{2}$ number of evidence factor nodes (solid square nodes) $\E^{\zeta=1}$; and $\frac{M}{2} \frac{N}{2}$ number of resolution dependency factor nodes (diagonal striped square) $\R^{\zeta=1}$. At each resolution level $\zeta$, labels of variable nodes $\V^{\zeta}$ represent the posterior disparity estimates; evidence factor nodes $\E^\zeta$ represent prior disparities; dependency factor nodes $\S^\zeta$ represent the likelihood of satisfying spatial dependencies given localized disparity estimates; and resolution dependency factor nodes $\R^{\zeta}$ represent the likelihood of agreement between disparity estimates between level $\zeta$ and level $\zeta - 1$ (multi-resolution dependencies).}
        \label{fig:MRFGS Disparity}       
    \end{figure}

The proposed multi-resolution factor graph model is a bipartite graph arranged as layers of factor graphs at multiple resolutions. Figure~\ref{fig:MRFGS Disparity} shows the schematic diagram of a two-level multi-resolution factor graph model for optimal estimation of disparities.

The multi-resolution bipartite graph is comprised of a set of variable nodes from multiple resolutions $\V = \collection{\V^\zeta}_{\zeta \in \Z_0^{+}}$ and a set of factor nodes from multiple resolutions $\F =  \collection{ \E^\zeta \cup \S^\zeta \cup \R^{\zeta+1}}_{\zeta \in \Z_0^{+}}$.  At each resolution level $\zeta$ with an associated scaling factor $\alpha = 2^\zeta$, the variable nodes $\V^\zeta = \collection{1, \ldots, \frac{M}{\alpha} \frac{N}{\alpha}}$ represent disparity labels assigned to each pixel at resolution level $\zeta$. Evidence factor nodes $\E^\zeta = \collection{1, \ldots, \frac{M}{\alpha} \frac{N}{\alpha}}$ provide prior degree of belief or evidence in assigning possible disparity labels at each pixel location at resolution level $\zeta$.  Dependency factor nodes $\S^\zeta = \collection{1, \ldots, \frac{M}{\alpha} \frac{N}{\alpha}}$ are used to model spatial dependencies among the disparity labels assigned to neighboring pixels within each resolution level $\zeta$.  Resolution dependency factor nodes $\R^{\zeta} = \collection{1, \ldots, \frac{M}{\alpha} \frac{N}{\alpha}}$ are used to model dependencies among the disparity labels $\D^{\zeta}$ at level $\zeta$ and $\D^{\zeta - 1}$ at level $\zeta -1$ (disparity labels $\D^{\zeta}$ formally defined below). There are no resolution dependency factor nodes in the MR-FGS graph when there is only one resolution level in the model (single resolution MR-FGS graph is same as the FGS graph).

At each resolution $\zeta$, a random field $\D^{\zeta} = \collection{d_i^\zeta: d_i^\zeta \in [d_{\min}^\zeta, d_{\max}^\zeta ] \subset \Z}_{i=1}^{\frac{M}{\alpha} \frac{N}{\alpha}}$ is assigned to variables nodes $\V^{\zeta}$.  A random variable $d_i^\zeta \in \D^\zeta$ is assigned to each \textit{variable node} $i \in \V^\zeta$ to represent the disparity label assigned to the $i$th pixel location at the multi-resolution level $\zeta$.  Within each level $\zeta$, each $j$th \textit{evidence factor node} in $\E^\zeta$ is connected one-to-one with the corresponding $i$th variable node in $\V^\zeta$ to incorporate prior belief or evidence in determining the disparity label $d_i^\zeta$.  Influence of each pixel location on its neighboring pixels is represented by connecting each variable node $i \in \V^\zeta$ with one or more \textit{dependency factor nodes} $\collection{k\in \S^\zeta}$ within each level $\zeta$ using local image characteristics as in the FGS model \cite{shabanian2021novel}. In brief, an $\alpha$th percentile cut-off of the \textit{bilateral filter} \cite{tomasi1998bilateral} coefficients estimated at the $i$th pixel location were used to identify neighboring variable nodes that have the highest influence on the true state of the disparity $d_i$.  To represent resolution dependency of pixel locations, each variable node $i \in \V^\zeta$ is connected to \textit{spatially corresponding} resolution dependency factor nodes in $R^{\zeta}$ and in $R^{\zeta - 1}$.

\subsection{MR-FGS Probabilistic Model}
Let, $ne(i)$ represent the set of neighboring nodes connected with any given node $i$; $ne(i)\setminus j$ represent a set of all neighboring nodes of $i$ excluding node $j$; $\D_j = \collection{d_i, \forall i \in \V : i \in ne\fof{j \in \F}}$ be a collection of random variables associated with any factor node $j \in \F$; and let, $\D_j\setminus i \subseteq \D$ be a collection of random variables in $\D_j$ except $d_i$.  

An \textit{evidence potential function} $\psi_j$ associated with each factor node $j \in \E^\zeta$ is defined as a function of random variables $\D_j$ of its neighboring nodes.  Because only one variable node is connected to an evidence factor node, $\psi_j\fof{\D_j} = \psi_j\fof{d_j}$.  As in the FGS model, we defined the evidence potential function $\psi_j\fof{d_j}$ as the \textit{a priori} distribution $p\fof{d_j}$.  A detailed description of estimating \textit{a priori} distributions at each resolution $\zeta$ is given in Section \ref{sec:a priori_MRFGS}.  Similarly, a \textit{spatial dependency potential function} $\psi_k(\D_k)$ of the $k$th dependency factor node $k \in \S^\zeta$ is defined as a function of the random variables $\D_k$ associated with its neighboring nodes $ne\fof{k}$. The potential function $\psi_k\fof{\D_k}$ was assigned a value of 1.0 when the states $\D_k \setminus i$ are same as that of $d_i$ and was assigned a value of $0$ otherwise.  The \textit{resolution dependency potential function} $\psi_r(\D_r)$ of the resolution dependency factor node $r \in \R^{\zeta}$ is defined as a function of the random variables $\D_r$ associated with its neighboring variable nodes $ne\fof{r}$ from multi-resolution levels $\zeta$ and $\zeta - 1$. The potential function $\psi_r\fof{\D_r}$ of a resolution dependency factor node $r \in \R^{\zeta}$ is assigned a value of 1.0 when the disparity labels of associated variable nodes in $\V^{\zeta - 1}$ is twice those of the associated variable nodes in $\V^{\zeta}$.  Otherwise, $\psi_r\fof{\D_r}$ was assigned 0.

The MR-FGS factor graph, therefore, represents joint distribution of disparity labels assigned to each of the pixel locations at all chosen resolutions in the model. The joint distribution of disparity labels of a model with two resolution levels, as shown in Figure~\ref{fig:MRFGS Disparity}, is as follows.
    \begin{align}
        p\fof{\D} &= \frac{1}{Z} \prod_{k \in \F} \psi_k\fof{\D_k}& \nonumber\\
            &=\frac{1}{Z} \prod_{j \in \E^{\zeta=0}} \psi_j\fof{d_j} \prod_{k \in \S^{\zeta=0}} \psi_k\fof{\D_k} \prod_{r \in \R^{\zeta=1}} \psi_r\fof{\D_r} \prod_{m \in \E^{\zeta=1}} \psi_m\fof{d_m} \prod_{n \in \S^{\zeta=1}} \psi_n\fof{\D_n} & \nonumber\\
            &=\frac{1}{Z}\left[ \prod_{\zeta=0}^{1} \prod_{j \in \E^{\zeta}} \psi_j\fof{d_j} \right] \left[ \prod_{\zeta=0}^{1} \prod_{k \in \S^{\zeta}} \psi_k\fof{\D_k} \right] \left[ \prod_{\zeta=1}^{1} \prod_{r \in \R^{\zeta}} \psi_r\fof{\D_r} \right]&\nonumber
    \end{align}
where, $Z$ is the partitioning function.  In general, for a model with $L$ multi-resolution levels, the joint distribution of the disparity labels assigned at all levels is
     \begin{align}
        p\fof{\D} &= \frac{1}{Z}\left[ \prod_{\zeta=0}^{L-1} \prod_{j \in \E^{\zeta}} \psi_j\fof{d_j} \right] \left[ \prod_{\zeta=0}^{L-1} \prod_{k \in \S^{\zeta}} \psi_k\fof{\D_k} \right] \left[ \prod_{\zeta=1}^{L-1} \prod_{r \in \R^{\zeta}} \psi_r\fof{\D_r} \right]&\label{eq:joint D}
     \end{align}
           
Probability of assigning various disparity labels to each pixel $i$ at any given multi-resolution level $\zeta$ can be obtained by marginalizing equation~\eqref{eq:joint D} with respect to $\D\setminus i$ as
    \begin{align} 
        p\fof{d_i} &= \frac{1}{Z} \sum_{\D\setminus d_i} \left[ \prod_{\zeta=0}^{L-1} \prod_{j \in \E^{\zeta}} \psi_j\fof{d_j} \right] \left[ \prod_{\zeta=0}^{L-1} \prod_{k \in \S^{\zeta=1}} \psi_k\fof{\D_k} \right] \left[ \prod_{\zeta=1}^{L-1} \prod_{r \in \R^{\zeta}} \psi_r\fof{\D_r} \right]&\label{eq:disp marginal_Multi}
    \end{align}

\subsection{Approximate Probabilistic Inference using Message Passing}
\label{sec:MR-FGS approximate MAP inference}
It can be observed that the \textit{sum-product} formulation in equation \eqref{eq:disp marginal_Multi} provides a posterior disparity estimate for each of the pixel locations $i$ at multi-resolution level $\zeta$ based on \textit{a priori} disparity information and spatial dependency characteristics of disparities at level $\zeta$ as well as based on the disparity labels of pixels (that are dependent on pixel $i$ at level $\zeta$) at levels $\zeta-1$ and $\zeta + 1$ \cite{pearl1982reverend,kschischang2001factor,shabanian2021novel}.  For an approximate and efficient computation of marginal beliefs or probabilities in equation~\eqref{eq:disp marginal_Multi} using \textit{loopy belief propagation}, local information available in each node is shared with neighboring nodes as variable-to-dependency factor messages $\mu_{i\in \V \rightarrow f \in (\S \cup \R) }$ and factor-to-variable messages $\mu_{f\in \F \rightarrow i\in \V}$ until convergence \cite{barber2012bayesian,murphy2013loopy}.  Each outgoing message from a node is defined as a function of incoming messages at the given node as follows \cite{barber2012bayesian}.
    \begin{align}
        \mu_{i\in \V \rightarrow f\in (\S \cup \R)} &= \prod_{g \in \collection{ne\fof{i}\setminus f}} \, \mu_{g \rightarrow i}\fof{d_i} & \label{eq:msg v-to-f_multi_resolution}\\
        \mu_{f\in \F \rightarrow i\in \V} &= \sum_{\D_f\setminus d_i} \, \psi_f\fof{\D_f} \prod_{j \in \collection{ne\fof{f}\setminus i}} \, \mu_{j\rightarrow f}\fof{d_j} & \label{eq:msg f-to-v_multi}
    \end{align}

The MR-FGS algorithm follows the same message passing structure as the FGS algorithm \cite{shabanian2021novel}. In brief, the variable-to-dependency factor node message $\mu_{i\in \V \rightarrow f\in \fof{\S \cup \R}}$ in equation~\eqref{eq:msg v-to-f_multi_resolution} approximated the posterior probability of $d_i$ while satisfying all of the constraints from its neighboring spatial and resolution dependency factor nodes except the factor node $f$ to which the message is sent.  Similarly, the factor node-to-variable node messages $\mu_{f\in \F \rightarrow i\in \V}$ in equation~\eqref{eq:msg f-to-v_multi} approximated the likelihood of satisfying spatial or resolution dependencies among the random variable states of all variables nodes associated with the factor node $f$ except $d_i$.  Relationship between the probabilistic solution structure of the MR-FGS algorithm with message passing is same as that of the FGS algorithm \cite{shabanian2021novel}.  Thus, the MR-FGS algorithm determines optimal disparities based on approximate inference of the posterior probability of assigning various disparity labels at each pixel $i$
    \begin{equation*}
        p\fof{d_i \mid \collection{f_j\fof{\D_j}, \forall j: i \in ne\fof{j}}} \propto p\fof{ \collection{f_j\fof{\D_j}, \forall j: i \in ne\fof{j}} \mid d_i} \, p\fof{d_i}
    \end{equation*}
where, $f_j\fof{\D_j}$ is a function of the state of all variable nodes neighboring the factor node $j$.

\subsection{Disparity Cost Volume and \textit{a Priori} Disparity Distribution $p\fof{d_i}$}
\label{sec:a priori_MRFGS}
Let, $C^\zeta\fof{i, d_i}$ be a cost volume representing the \textit{a priori} cost of assigning a disparity label $d_i$  to pixel location $i$ at resolution level $\zeta$.  The approximate \textit{maximum a posteriori} inference described in Section \ref{sec:MR-FGS approximate MAP inference} updates this \textit{a priori} cost volume based on the spatial and resolution dependencies defined by the model.  Thus, the MR-FGS disparity estimates form an optimal surface (optimal in \textit{maximum a posteriori} sense) within the posterior cost volume obtained after message convergence.  

\begin{figure}
    \includegraphics[width=1\textwidth,keepaspectratio]{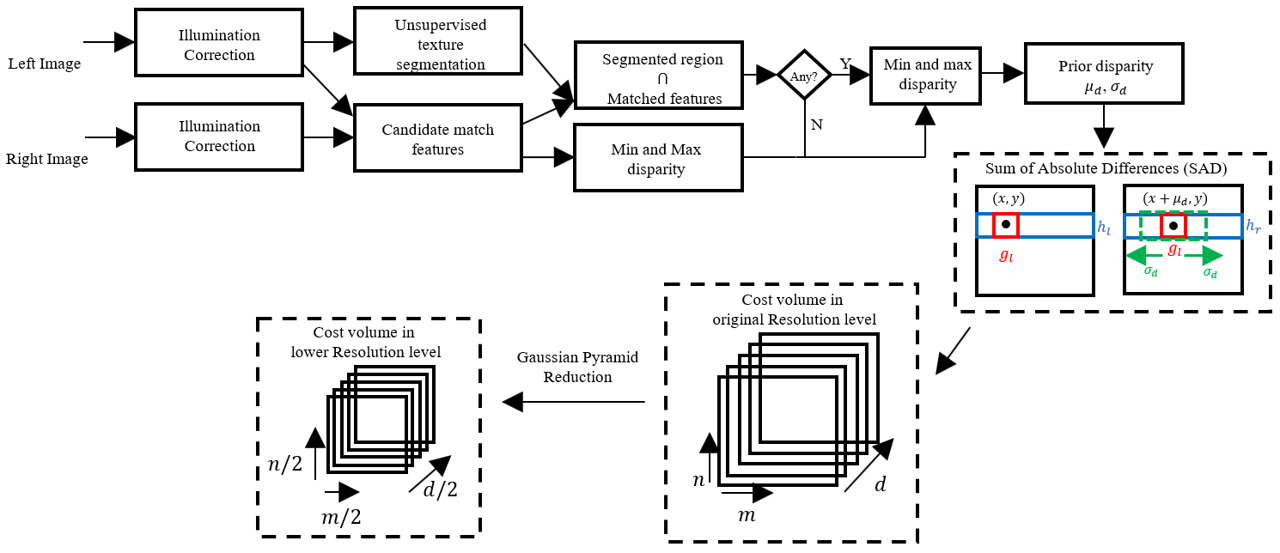}\centering
    \caption{Schematic representation of algorithmic steps used for computing \textit{a priori} cost volumes at resolutions $\zeta = 0$ and $\zeta = 1$.}
    \label{fig:zonal disparity_MRFGS}       
\end{figure}

Figure~\ref{fig:zonal disparity_MRFGS} shows the computational steps used for computing the \textit{a priori} cost volumes $C^{\zeta=0,1}\fof{i, d_i}$ at resolutions $\zeta = 0$ and $\zeta = 1$. In brief, one of the images in each stereo pair (reference image) at resolution $\zeta = 0$ was segmented using the the Gabor unsupervised texture segmentation algorithm \cite{jain1991unsupervised}.  For each segmented region in the stereo pair, highly confident sparse stereo coordinate correspondences were identified using an eigen-based feature matching method \cite{shi1994good} and the estimated sparse disparities were assumed to be normally distributed $N\fof{\mu_d, \sigma_d}$.  Using the zonal distribution of disparities, the cost of assigning disparities $d_i \in \left[\mu_d - \sigma_d, \mu_d + \sigma_d\right]$ to each of the pixels $i$ within the segmented region were estimated.  Sum of absolute differences (SAD) was used as the cost of associating a disparity label $d_i$ to the $i$th pixel.  A sparse cost volume $C{^\zeta}$ at any resolution $\zeta$ was computed by downsampling the cost volume $C{^{\zeta-1}}$ from the next higher resolution $\fof{\zeta-1}$ by 2 after low-pass filtering \cite{burt1983edward}. \textit{A priori} probability $p^{\zeta}\fof{d_i}$ of assigning a disparity label $d_i$ to the $i$th pixel location at resolution $\zeta$ was estimated using the corresponding cost volume $C^{\zeta}\fof{i, d_i}$ as follows.
  \begin{equation}
            p^{\zeta}\fof{d_i}=\frac{C^{\zeta}\fof{i, d_i}}{\sum_{d_j} C{^\zeta}\fof{i, d_j}}
             \label{eq:disp_marginal} 
        \end{equation}

\SetNlSty{textbf}{}{:}
\SetNlSkip{1.25em}
\RestyleAlgo{ruled}
\IncMargin{1em}
\begin{algorithm}
    \SetKwData{Left}{left}\SetKwData{This}{this}\SetKwData{Up}{up}
    \SetKwFunction{Union}{Union}\SetKwFunction{FindCompress}{FindCompress}
    \SetKwInOut{Input}{Input}\SetKwInOut{Output}{Output}
    \Input{A sparse disparity cost volume $C(i,d_i)$ from a stereo pair of size $M \times N$ pixels and 
            convergence threshold $\tau$.}
    \Output{An optimal disparity estimate at each resolution $\zeta$: \; $\D^\zeta = \collection{d_i^\zeta: d_i^\zeta \in [d_{\min}^\zeta, d_{\max}^\zeta ] \subset \Z : \alpha = 2^\zeta}_{i=1}^{\frac{M}{\alpha} \frac{N}{\alpha}}$.}
    \BlankLine
    Compute \textit{a down-sampled} sparse cost volume for the lower resolution levels: $  C^\zeta(i,d_i)$\;

    Estimate \textit{a priori} probability from cost volume at each resolution levels: $p^\zeta(d_i) \leftarrow \frac{C^\zeta(i, d_i)}{\sum_{d_j} C^\zeta(i, d_j)}$\;
    %
    Define factor graph nodes namely variable nodes $\V^\zeta$, evidence factor nodes $\E^\zeta$, dependency factor nodes $\S^\zeta$, and resolution factor nodes $\R^{\zeta - 1}$ with $\F =  \collection{ \E^\zeta \cup \S^\zeta \cup \R^{\zeta-1}}_{\zeta \in \Z_0^{+}}$\;
    Form factor graph in multi resolution levels by connecting each evidence factor node $j \in \E^\zeta$ with each variable node $i \in \V^\zeta$; and each variable node $i \in \V^\zeta$ with one or more dependency factor nodes $k \in \S^\zeta$  at resolution level $\zeta$; each $r$th \textit{resolution factor node} in $\R^{\zeta-1}$ is connected to corresponding variable nodes $i^{\zeta} \in \V^\zeta$ of the original resolution level; each variable node $i \in \V^{\zeta+1}$ is connected to \textit{spatially corresponding} resolution dependency factor nodes in $\R^\zeta$ and in $\R^{\zeta-1}$\;
    Associate each variable node $i$ with a random variable $d_i$ to represent disparity label at the $i$th pixel location\;
    Initialize evidence factor nodes in each resolution $\zeta$ with the prior probabilities: $j \in \E^\zeta \leftarrow  p^\zeta(d_i)$, where $ i \in ne(j)$\;    
    Define an evidence potential function $\psi_j, j \in E^\zeta$ as a function of random variables $\D_j$ of its neighboring nodes $\psi_j(\D_j)=\psi_j(d_i)$\;
    Define a spatial dependency potential function $\psi_k, k \in \S^\zeta$ as a function of random variables $\D_k$ of its neighboring nodes $\psi_k(\D_k)$\;
    Define a resolution potential function $\psi_r, r \in \R$ as a function of random variables $\D_r$ of its neighboring nodes $\psi_r(\D_r)$\;
    \While(Loopy belief propagation){convergence error $\epsilon^{t+1}$ in \cite{shabanian2021novel} $>$ threshold $\tau$}{
    \tab Send variable-to-dependency-factor messages: $\mu_{i\in \V \rightarrow f\in \S }$ as in \eqref{eq:msg v-to-f_multi_resolution}\;
    \tab Send factor-to-variable messages: $\mu_{f\in \F \rightarrow i\in \V}$ as in \eqref{eq:msg f-to-v_multi}\;
    }
    Determine \textit{maximum a posteriori} disparity estimate $\hat{d}_i$ at each pixel in the original resolution level: $\hat{d}_i = \argmax_{d_i} p\fof{d_i \mid \collection{f_j\fof{\D_j}, \forall j: i \in ne\fof{j}}}$\;
    Update disparity estimates in occluded regions using \textit{weighted median filter} as in Section~\ref{sec:postprocessing_MRFGS}\;
    \caption{Probabilistic multi-resolution factor graph-based disparity estimation (MR-FGS) Algorithm}\label{algo_disjdecomp_MRFGS}
    \label{alg:fgs_MRFGS}
\end{algorithm}\DecMargin{1em}
\subsection{Post-processing}
\label{sec:postprocessing_MRFGS}
In general, stereo matching algorithms including the FGS algorithm \cite{shabanian2021novel} utilize a \textit{left-right consistency} check to improve the accuracy of the disparity estimates specifically in the regions with pixel occlusions.  Multi-resolution strategy used in the MR-FGS algorithm eliminated the need for such left-right consistency checks and significantly reduced the computing time. \textit{Maximum a posterior} disparity estimates were filtered using a  weighted median filter to further improve the estimation accuracy \cite{brownrigg1984weighted}.

%
\section{Experimental Results and Discussion}
\label{Experiments_MRFGS}
Algorithm~\ref{alg:fgs_MRFGS} depicts detailed computational steps of the MR-FGS algorithm. Performance of the MR-FGS was evaluated and compared with the FGS algorithm using rectified stereo images from four Middlebury benchmark datasets \cite{scharstein2003high, scharstein2007learning, hirschmuller2007evaluation, scharstein2014high}.  Similar to the FGS algorithm, in most of the experiments, the MR-FGS algorithm converged between 20-30 iterations based on an $L_2$ measure of change in disparity estimates between successive iterations.
\subsection{MR-FGS Parameters and Implementation}
The MR-FGS algorithm was implemented in MATLAB 2018b and evaluated using an Intel(R) Xeon(R) workstation with E3-1271 v3, 3.6 GHz processor. Illumination differences between images in each stereo pair were corrected using a homomorphic filter of size $21 \times 21$. The Gabor filter was used to perform texture segmentation with filter orientations between $0\degree-150\degree$ degrees in steps of $30\degree$, wavelength starting from $2.83$ up to the magnitude of hypotenuse of the input image.  $K$-means clustering algorithm was initialized with $K=15$ with $5$ replicates and ran for $500$ iterations. At each pixel location, SAD cost was computed using a window size of $7 \times 7$ pixels. Bilateral filters with a kernel size of $7 \times 7$ pixels, domain kernel parameter of $\sigma_d = 3$, range kernel parameter of $\sigma_r = 0.1$ and coefficient percentile cut-off of $\alpha=97$ were used to identify neighboring pixels with significant influence on any $i$th pixel in the model. 
%
\subsection{Performance Metrics}
Accuracy of the MR-FGS algorithm in estimating the stereo correspondences was assessed qualitatively based on the \textit{disparity error maps} and quantitatively using the performance metrics of \textit{average absolute error} (Avg. err in pixels) and peak signal-to-noise ratio (PSNR). Disparity error maps were computed as location-wise difference between the estimated disparity  $\hat{d}(x,y)$ and its ground-truth $d(x,y)$ as $\hat{d}(x,y) - d(x,y)$.  PSNR provides a measure of similarity between an estimated disparity map $\hat{d}(x,y)$ of size $M \times N$ pixels and the ground-truth disparity map $d(x,y)$ as follows.
    \begin{equation}
        PSNR=10 \log _{10}\frac{255^2 \times M \times N}{\sum_{\forall(x,y)}\fof{\hat{d}(x,y)-d(x,y)}^2}
    \end{equation}
A thresholded average disparity error metric with a disparity threshold of $T$ pixels was defined as  
    \begin{equation}
         Bad=\fof{\frac{1}{MN}\sum_{\forall(x,y)}(|\hat{d}(x,y)- d(x,y)|>T)}\times 100
    \end{equation}
Average disparity errors were assessed at two disparity threshold levels of $T=2$ pixels (Bad2.0) and  $T=0$ pixels (Avg. err).

\subsection{Performance Assessment of the MR-FGS Algorithm using Benchmark Datasets}
Five Middlebury stereo datasets with differing illumination and textures namely Teddy and Cones from 2003 \cite{scharstein2003high}, Dolls stereo pair from 2005 \cite{scharstein2007learning}, Rocks1 from 2006 \cite{hirschmuller2007evaluation}, and Motorcycle from 2014 \cite{scharstein2014high} were used to evaluate the performance of MR-FGS algorithm and to compare with the FGS algorithm.
\subsubsection{Accuracy of Disparity Maps Without Post-processing:}
\begin{figure}[!htb]
\centering
   \rotatebox[origin=c]{90}{\tiny Ground truth}
     \begin{subfigure}[a]{0.19\textwidth}
        \centering
         \includegraphics[width=\textwidth]{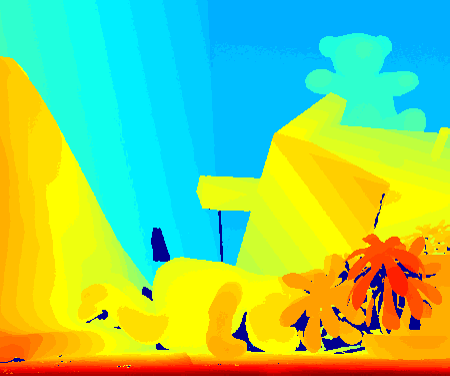}
         \label{fig:3_MRFGS.a.1}
     \end{subfigure}
     \hfill
     \begin{subfigure}[a]{0.19\textwidth}
         \centering
         \includegraphics[width=\textwidth]{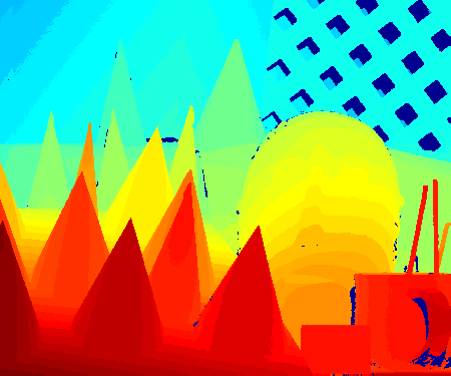}
         \label{fig:3_MRFGS.b.1}
     \end{subfigure} 
     \hfill
      \begin{subfigure}[a]{0.19\textwidth}
         \centering
         \includegraphics[width=\textwidth]{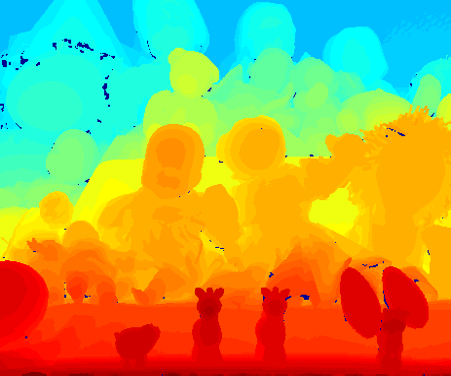}
         \label{fig:3_MRFGS.c.1}
     \end{subfigure}
      \hfill
      \begin{subfigure}[a]{0.19\textwidth}
         \includegraphics[width=\textwidth]{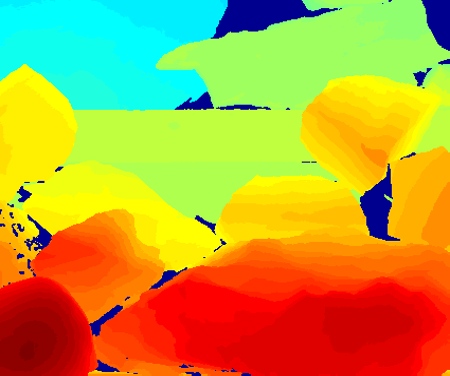}
         \label{fig:3_MRFGS.d.1}
     \end{subfigure}     
     \hfill
      \begin{subfigure}[a]{0.19\textwidth}
         \includegraphics[width=\textwidth]{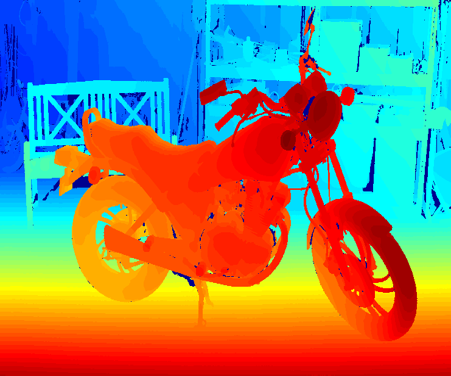}
         \label{fig:3_MRFGS.e.1}
     \end{subfigure}  \\   
  
   \centering \rotatebox[origin=c]{90}{\tiny FGS disparity}
      \begin{subfigure}[a]{0.19\textwidth}
         \centering
         \includegraphics[width=\textwidth]{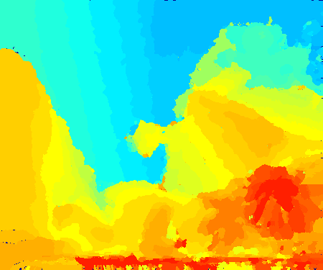}
         \label{fig:3_MRFGS.a.2}
     \end{subfigure}
     \hfill
     \begin{subfigure}[a]{0.19\textwidth}
         \centering
         \includegraphics[width=\textwidth]{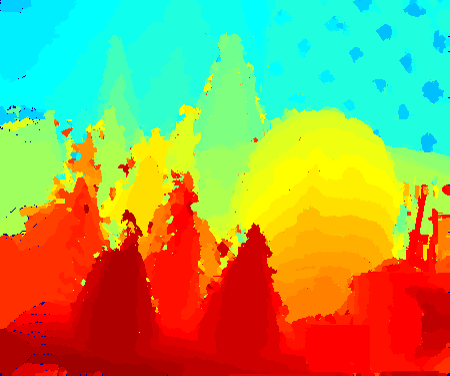}
         \label{fig:3_MRFGS.b.2}
     \end{subfigure} 
     \hfill
      \begin{subfigure}[a]{0.19\textwidth}
         \centering
         \includegraphics[width=\textwidth]{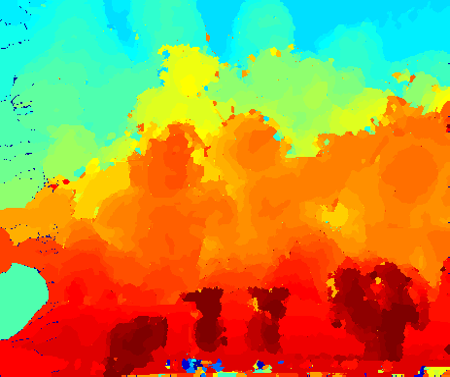}
         \label{fig:3_MRFGS.c.2}
     \end{subfigure}
      \hfill
      \begin{subfigure}[a]{0.19\textwidth}
         \includegraphics[width=\textwidth]{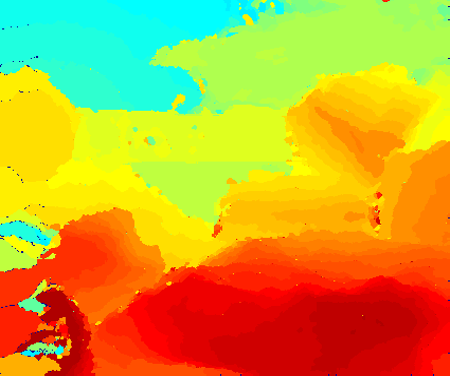}
         \label{fig:3_MRFGS.d.2}
     \end{subfigure}     
     \hfill
      \begin{subfigure}[a]{0.19\textwidth}
         \includegraphics[width=\textwidth]{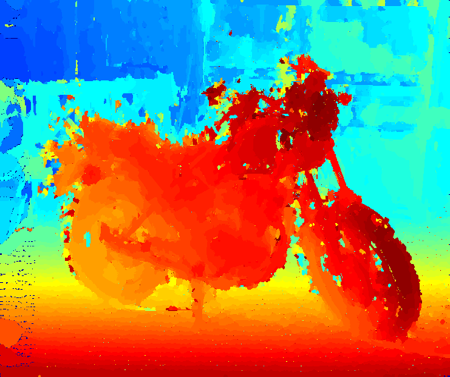}
         \label{fig:3_MRFGS.e.2}
     \end{subfigure}  \\   
   
      
      \rotatebox[origin=c]{90}{\tiny \tab FGS error map}
      \begin{subfigure}[a]{0.19\textwidth}
         \centering
         \includegraphics[width=\textwidth]{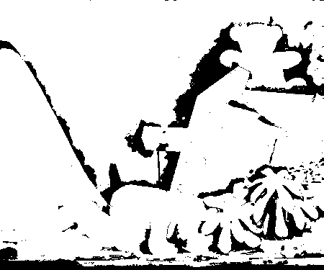}
         \label{fig:3_MRFGS.a.3} 
     \end{subfigure}
     \hfill
     \begin{subfigure}[a]{0.19\textwidth}
         \centering
         \includegraphics[width=\textwidth]{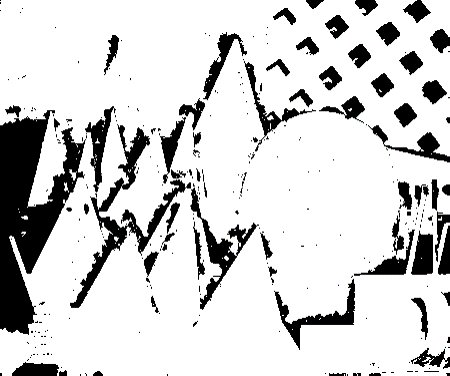}
         \label{fig:3_MRFGS.b.3} 
     \end{subfigure} 
     \hfill
      \begin{subfigure}[a]{0.19\textwidth}
         \centering
         \includegraphics[width=\textwidth]{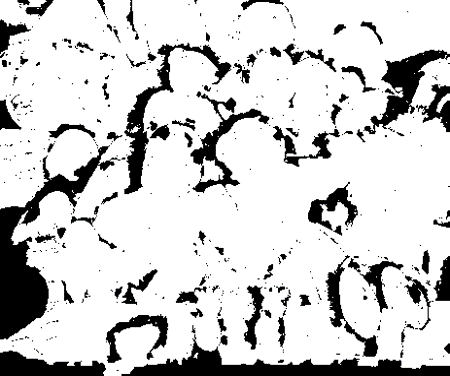}
         \label{fig:3_MRFGS.c.3} 
     \end{subfigure}
      \hfill
      \begin{subfigure}[a]{0.19\textwidth}
         \includegraphics[width=\textwidth]{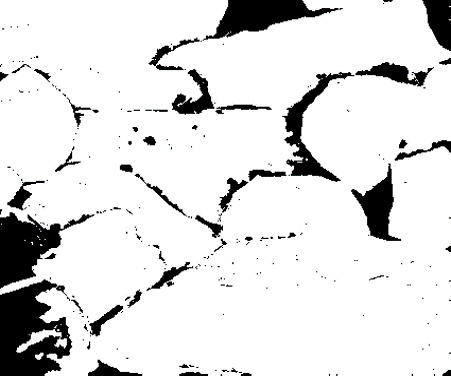}
         \label{fig:3_MRFGS.d.3}
     \end{subfigure}     
     \hfill
      \begin{subfigure}[a]{0.19\textwidth}
         \includegraphics[width=\textwidth]{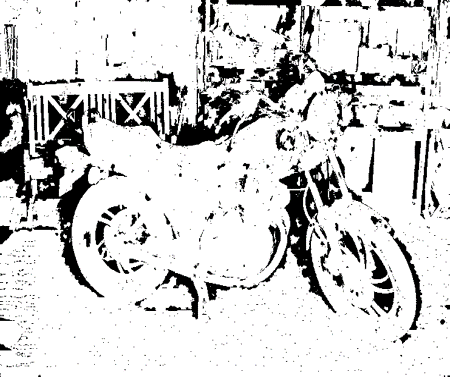}
         \label{fig:3_MRFGS.e.3} 
     \end{subfigure}  \\   

     \centering \rotatebox[origin=c]{90}{\tiny MR-FGS disparity}
      \begin{subfigure}[a]{0.19\textwidth}
         \centering
         \includegraphics[width=\textwidth]{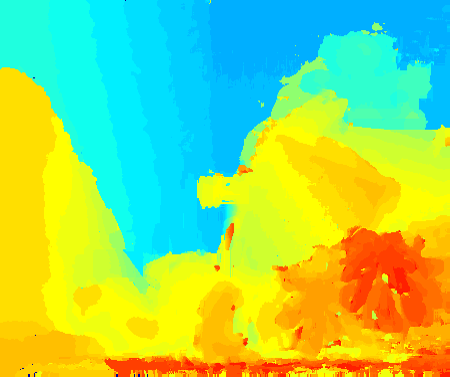}
         \label{fig:3_MRFGS.a.4}
     \end{subfigure}
     \hfill
     \begin{subfigure}[a]{0.19\textwidth}
         \centering
         \includegraphics[width=\textwidth]{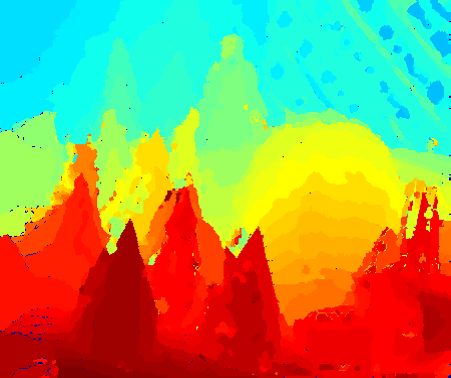}
         \label{fig:3_MRFGS.b.4}
     \end{subfigure} 
     \hfill
      \begin{subfigure}[a]{0.19\textwidth}
         \centering
         \includegraphics[width=\textwidth]{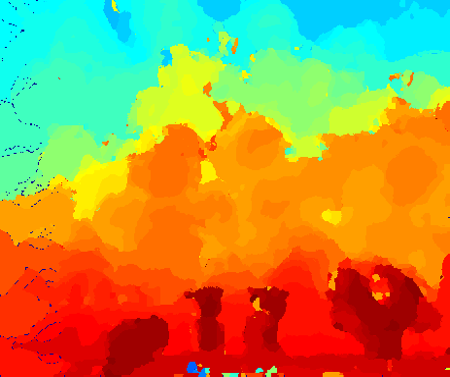}
         \label{fig:3_MRFGS.c.4}
     \end{subfigure}
      \hfill
      \begin{subfigure}[a]{0.19\textwidth}
         \includegraphics[width=\textwidth]{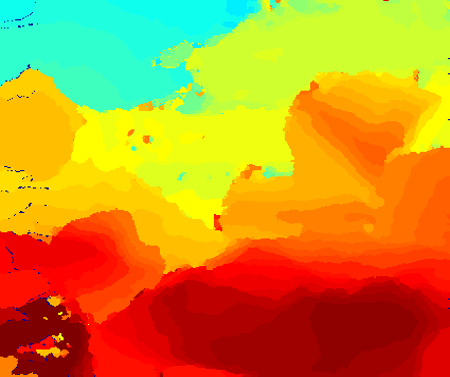}
         \label{fig:3_MRFGS.d.4}
     \end{subfigure}     
     \hfill
      \begin{subfigure}[a]{0.19\textwidth}
         \includegraphics[width=\textwidth]{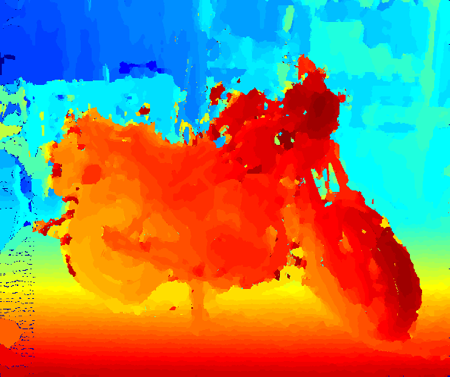}
         \label{fig:3_MRFGS.e.4}
     \end{subfigure}  \\   
      
      \rotatebox[origin=c]{90}{\tiny \tab MR-FGS error map}
      \begin{subfigure}[a]{0.19\textwidth}
         \centering
         \includegraphics[width=\textwidth]{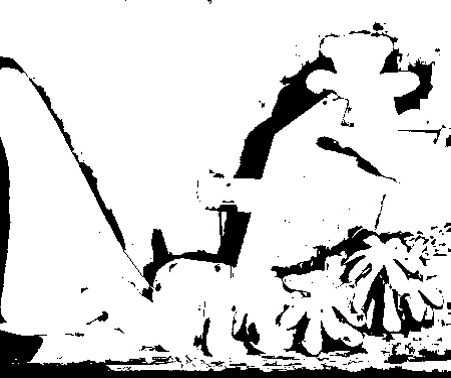}
         \label{fig:3_MRFGS.a.5} \caption{}
     \end{subfigure}
     \hfill
     \begin{subfigure}[a]{0.19\textwidth}
         \centering
         \includegraphics[width=\textwidth]{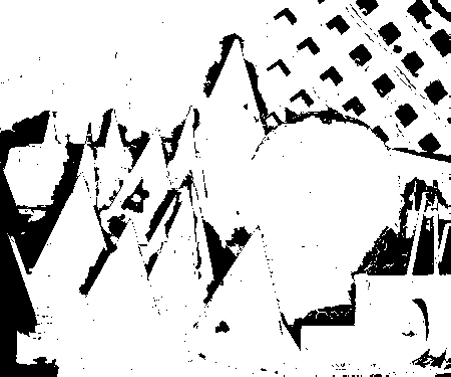}
         \label{fig:3_MRFGS.b.5} \caption{}
     \end{subfigure} 
     \hfill
      \begin{subfigure}[a]{0.19\textwidth}
         \centering
         \includegraphics[width=\textwidth]{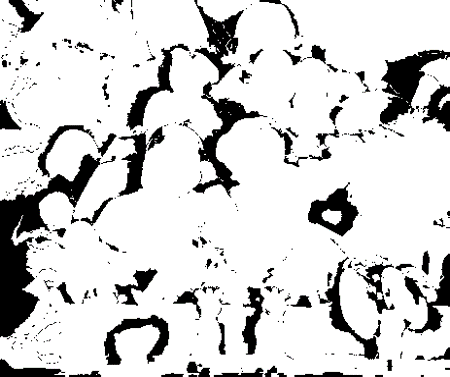}
         \label{fig:3_MRFGS.c.5} \caption{}
     \end{subfigure}
      \hfill
      \begin{subfigure}[a]{0.19\textwidth}
         \includegraphics[width=\textwidth]{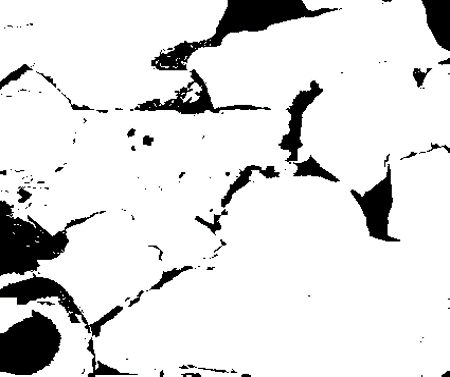}
         \label{fig:3_MRFGS.d.5} \caption{}
     \end{subfigure}     
     \hfill
      \begin{subfigure}[a]{0.19\textwidth}
         \includegraphics[width=\textwidth]{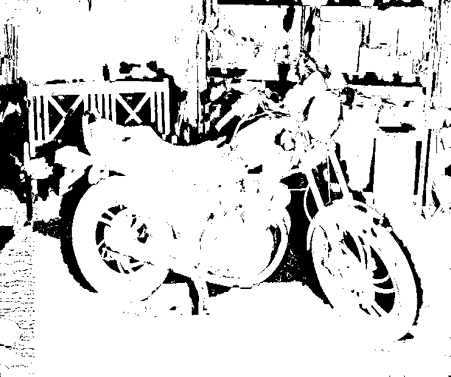}
         \label{fig:3_MRFGS.e.5} \caption{}
     \end{subfigure}  \\   
    \caption{Disparity estimates and error maps of MR-FGS and FGS without any post-processing. For assessing the disparity estimates, 5 stereo pairs were selected from the Middlebury benchmark dataset namely (a) Teddy (dataset 2003), (b) Cones (dataset 2003), (c) Dolls (dataset 2005), (d) Rocks1 (dataset 2006), and (e) Motorcycle (dataset 2014). Occluded regions were not excluded in the disparity estimates.  For all the datasets, the MR-FGS algorithm provided more sharper depth boundaries than the FGS algorithm.}
    \label{fig:3_MRFGS}       
\end{figure}

Figure~\ref{fig:3_MRFGS} shows the ground truth disparity maps, estimated disparity maps without any post-processing and corresponding disparity error maps for the MR-FGS and FGS algorithms.  The MR-FGS algorithm provided sharper depth boundaries than the FGS algorithm.  Table~\ref{tab:1_MRFGS} presents quantitative performance metrics without post-processing the FGS and MR-FGS disparity estimates.  For all datasets, the MR-FGS method provided higher accuracy by all performance metrics than the FGS algorithm.
%
\begin{table}[!htb]
    \centering
    \caption{Performance evaluation of the MR-FGS and FGS algorithms on Middlebury stereo pairs without performing any post-processing. The occluded regions were included in the performance evaluations. For all datasets, the MR-FGS algorithm provided more accurate disparity estimates than the FGS algorithm based on all performance metrics.}
    \label{tab:1_MRFGS}       
    \begin{tabular}{cccccccc}
    \hline\noalign{\smallskip}
        Images & \multicolumn{3}{c}{FGS \cite{shabanian2021novel}} & & \multicolumn{3}{c}{MR-FGS Algorithm}\\
        \cline{2-4}\cline{6-8}\noalign{\smallskip} 
        &Avg.err & PSNR(dB) & Bad2.0(\%) & & Avg.err & PSNR(dB) & Bad2.0(\%)\\
        \noalign{\smallskip}\hline\noalign{\smallskip}
        Teddy      &  2.60 & 32.02 & 14.17 && 2.09 & 32.43 & 13.90 \\
        Cones      &  2.78 & 32.35 & 17.60 &&  2.29 & 35.14 & 16.52 \\
        Dolls      & 3.20 & 30.75 & 22.47 && 2.63 &35.11 & 18.50 \\
        Rocks1     &  3.29 & 30.15 & 13.58 && 3.21 & 33.31 & 11.96 \\
        Motorcycle & 3.81 & 29.45 & 20.04 && 3.40 & 29.90 & 19.53 \\
    \noalign{\smallskip}\hline\noalign{\smallskip}
    \end{tabular}
\end{table}

\clearpage

\subsubsection{Accuracy of Disparity Maps After Post-processing:} 
\begin{figure}[hbt!]
\centering
     \rotatebox[origin=c]{90}{\tiny Ground truth}
     \begin{subfigure}[a]{0.19\textwidth}
         \centering
         \includegraphics[width=\textwidth]{Figures/Teddy_GT.png}
         \label{fig:4_MRFGS.a.1}
     \end{subfigure}
     \hfill
     \begin{subfigure}[a]{0.19\textwidth}
         \centering
         \includegraphics[width=\textwidth]{Figures/Cones_GT.png}
         \label{fig:4_MRFGS.b.1}
     \end{subfigure} 
     \hfill
      \begin{subfigure}[a]{0.19\textwidth}
         \centering
         \includegraphics[width=\textwidth]{Figures/Dolls_GT.png}
         \label{fig:4_MRFGS.c.1}
     \end{subfigure}
      \hfill
      \begin{subfigure}[a]{0.19\textwidth}
         \includegraphics[width=\textwidth]{Figures/Rocks1_GT.png}
         \label{fig:4_MRFGS.d.1}
     \end{subfigure}     
     \hfill
      \begin{subfigure}[a]{0.19\textwidth}
         \includegraphics[width=\textwidth]{Figures/Motorcycle_GT.png}
         \label{fig:4_MRFGS.e.1}
     \end{subfigure}  \\ 
         \rotatebox[origin=c]{90}{\tiny \hspace{0.5cm} FGS final disparity}
     \begin{subfigure}[a]{0.19\textwidth}
         \centering
         \includegraphics[width=\textwidth]{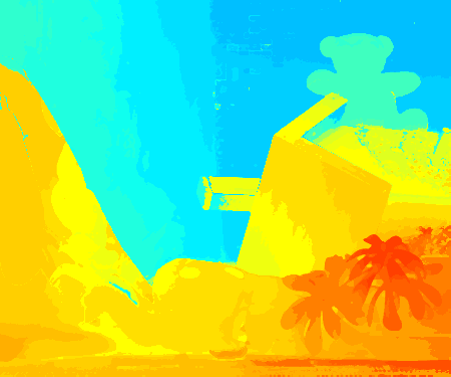}
         \label{fig:4_MRFGS.a.2}
     \end{subfigure}
     \hfill
     \begin{subfigure}[a]{0.19\textwidth}
         \centering
         \includegraphics[width=\textwidth]{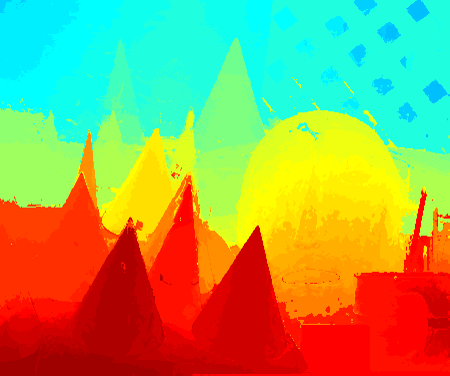}
         \label{fig:4_MRFGS.b.2}
     \end{subfigure} 
     \hfill
      \begin{subfigure}[a]{0.19\textwidth}
         \centering
         \includegraphics[width=\textwidth]{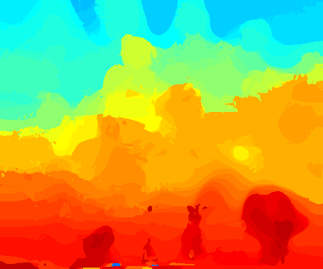}
         \label{fig:4_MRFGS.c.2}
     \end{subfigure}
      \hfill
      \begin{subfigure}[a]{0.19\textwidth}
         \includegraphics[width=\textwidth]{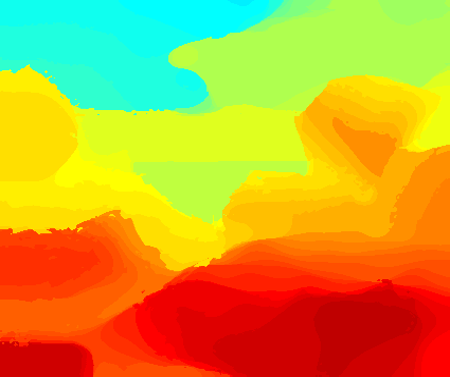}
         \label{fig:4_MRFGS.d.2}
     \end{subfigure}     
     \hfill
      \begin{subfigure}[a]{0.19\textwidth}
         \includegraphics[width=\textwidth]{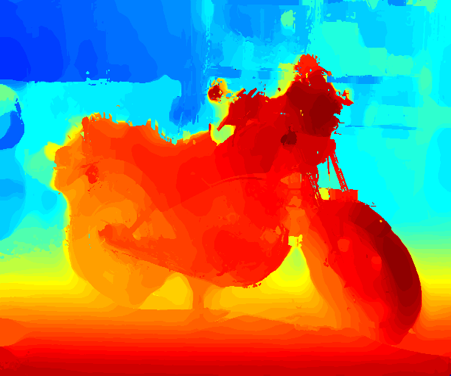}
         \label{fig:4_MRFGS.e.2}
     \end{subfigure}  \\   
      \rotatebox[origin=c]{90}{\tiny \tab FGS error map}        
    \begin{subfigure}[a]{0.19\textwidth}
         \centering
         \includegraphics[width=\textwidth]{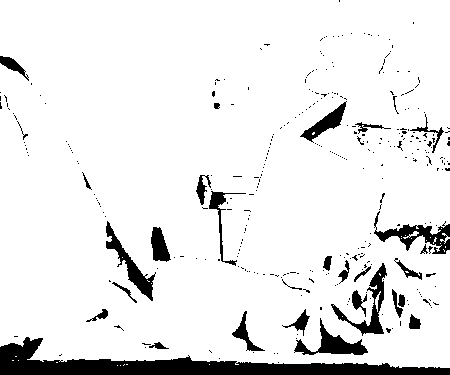}
         \label{fig:4_MRFGS.a.3} 
     \end{subfigure}
     \hfill
     \begin{subfigure}[a]{0.19\textwidth}
         \centering
         \includegraphics[width=\textwidth]{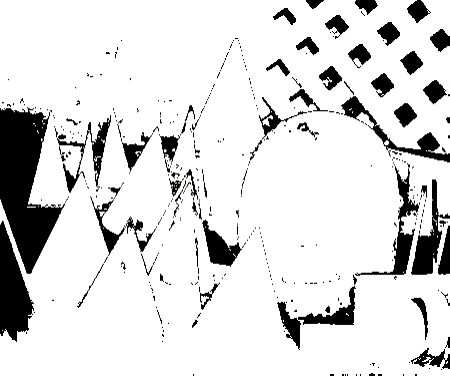}
         \label{fig:4_MRFGS.b.3} 
     \end{subfigure} 
     \hfill
      \begin{subfigure}[a]{0.19\textwidth}
         \centering
         \includegraphics[width=\textwidth]{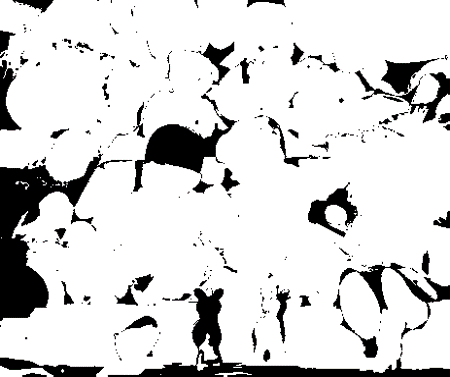}
         \label{fig:4_MRFGS.c.3}
     \end{subfigure}
      \hfill
      \begin{subfigure}[a]{0.19\textwidth}
         \includegraphics[width=\textwidth]{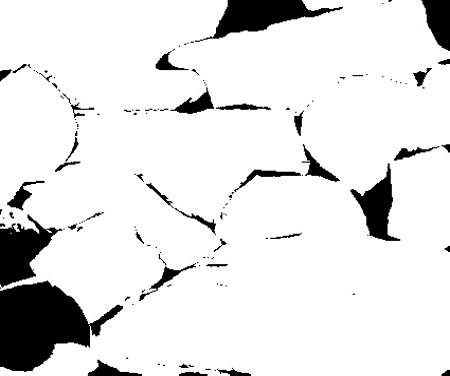}
         \label{fig:4_MRFGS.d.3}
     \end{subfigure}     
     \hfill
      \begin{subfigure}[a]{0.19\textwidth}
         \includegraphics[width=\textwidth]{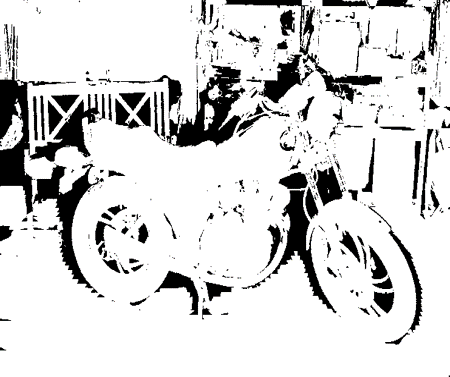}
         \label{fig:4_MRFGS.e.3}
     \end{subfigure}  \\ 
         \rotatebox[origin=c]{90}{\tiny \hspace{0.5cm} MR-FGS final disparity}
     \begin{subfigure}[a]{0.19\textwidth}
         \centering
         \includegraphics[width=\textwidth]{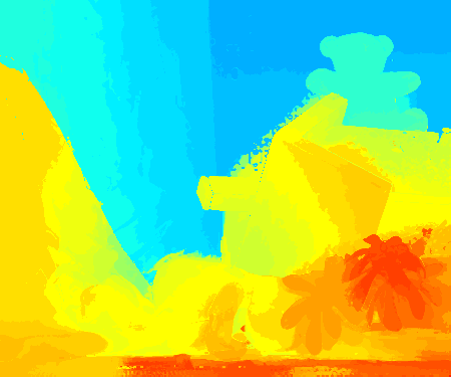}
         \label{fig:4_MRFGS.a.4}
     \end{subfigure}
     \hfill
     \begin{subfigure}[a]{0.19\textwidth}
         \centering
         \includegraphics[width=\textwidth]{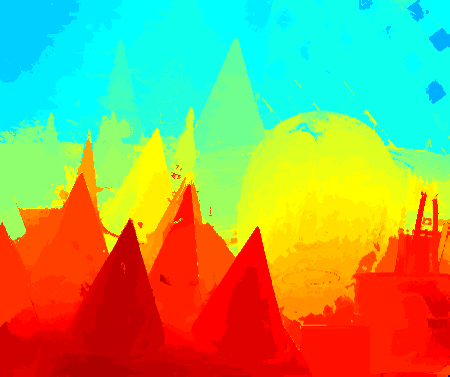}
         \label{fig:4_MRFGS.b.4}
     \end{subfigure} 
     \hfill
      \begin{subfigure}[a]{0.19\textwidth}
         \centering
         \includegraphics[width=\textwidth]{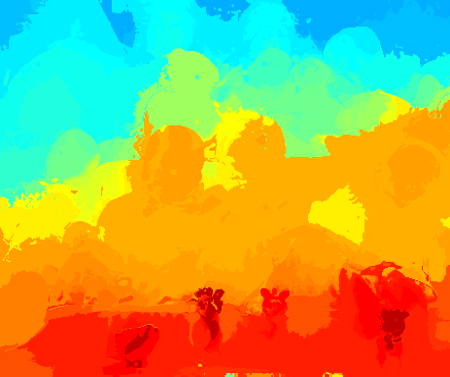}
         \label{fig:4_MRFGS.c.4}
     \end{subfigure}
      \hfill
      \begin{subfigure}[a]{0.19\textwidth}
         \includegraphics[width=\textwidth]{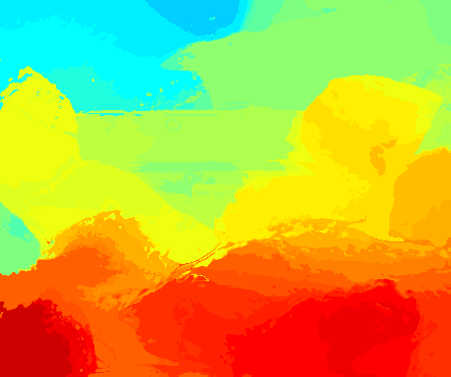}
         \label{fig:4_MRFGS.d.4}
     \end{subfigure}     
     \hfill
      \begin{subfigure}[a]{0.19\textwidth}
         \includegraphics[width=\textwidth]{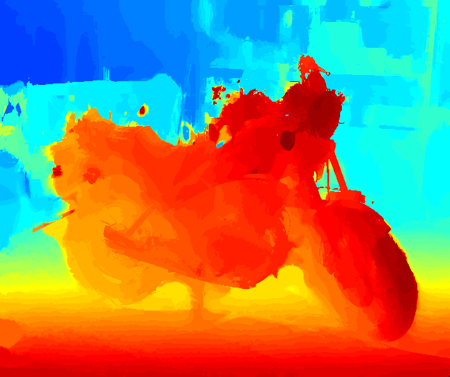}
         \label{fig:4_MRFGS.e.4}
     \end{subfigure}  \\   
      \rotatebox[origin=c]{90}{\tiny \tab MR-FGS error map}        
    \begin{subfigure}[a]{0.19\textwidth}
         \centering
         \includegraphics[width=\textwidth]{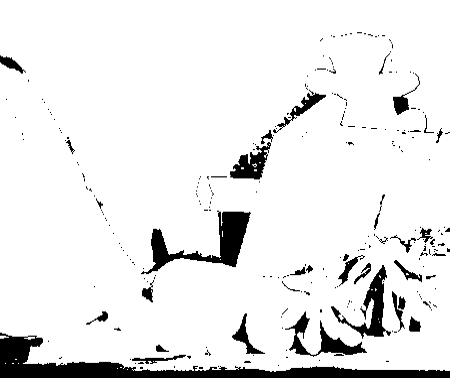}
         \label{fig:4_MRFGS.a.5} \caption{}
     \end{subfigure}
     \hfill
     \begin{subfigure}[a]{0.19\textwidth}
         \centering
         \includegraphics[width=\textwidth]{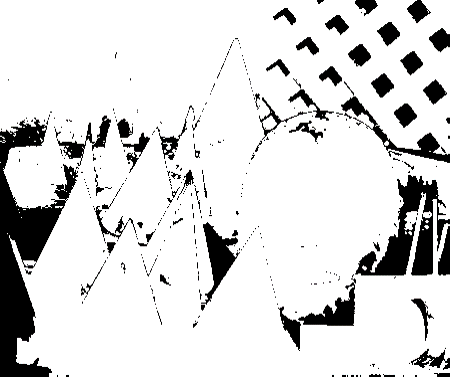}
         \label{fig:4_MRFGS.b.5} \caption{}
     \end{subfigure} 
     \hfill
      \begin{subfigure}[a]{0.19\textwidth}
         \centering
         \includegraphics[width=\textwidth]{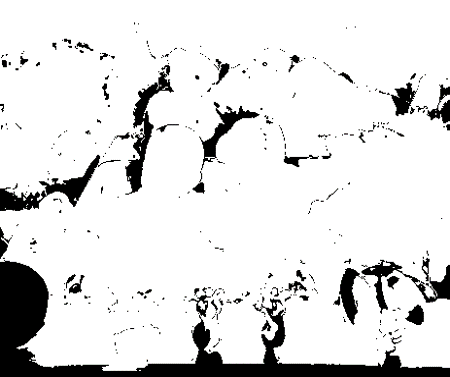}
         \label{fig:4_MRFGS.c.5} \caption{}
     \end{subfigure}
      \hfill
      \begin{subfigure}[a]{0.19\textwidth}
         \includegraphics[width=\textwidth]{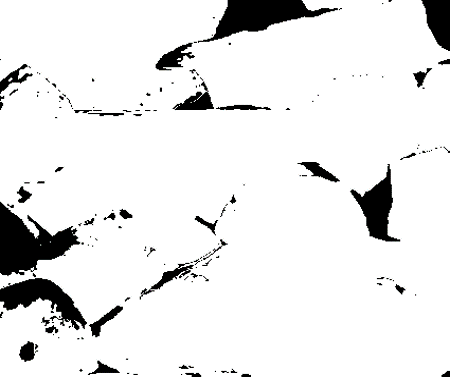}
         \label{fig:4_MRFGS.d.5} \caption{}
     \end{subfigure}     
     \hfill
      \begin{subfigure}[a]{0.19\textwidth}
         \includegraphics[width=\textwidth]{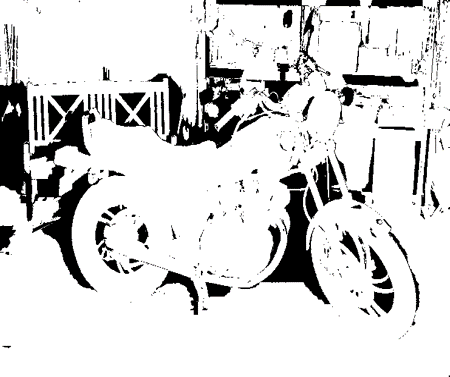}
         \label{fig:4_MRFGS.e.5} \caption{}
     \end{subfigure}  \\ 

    \caption{Disparity estimates and error maps of MR-FGS and FGS after post-processing. FGS algorithm requires additional computationally expensive left-right consistency checks.  Performance evaluated using 5 stereo pairs from the Middlebury benchmark dataset namely (a) Teddy (dataset 2003), (b) Cones (dataset 2003), (c) Dolls (dataset 2005), (d) Rocks1 (dataset 2006), and (e) Motorcycle (dataset 2014). Occluded regions were not excluded in the disparity estimates.  MR-FGS disparity estimates were more accurate than those of the FGS algorithm.}
    \label{fig:4_MRFGS}       
\end{figure}

Figure~\ref{fig:4_MRFGS} shows the ground truth disparity maps, estimated disparity maps after post-processing and corresponding disparity error maps for the MR-FGS and FGS algorithms.  When compared with the ground truth disparity maps, MR-FGS disparity estimates were more accurate than those of the FGS algorithm.  Table~\ref{tab:2} presents quantitative performance metrics after post-processing the FGS and MR-FGS disparity estimates.  For all datasets, accuracy of the MR-FGS disparity estimates were higher for all datasets than the FGS disparity estimates.  In contrast to the FGS algorithm, MR-FGS algorithm does not require computationally expensive left-right consistency checks for the disparity estimates.

\begin{table}[hbt!]
    \centering
    \caption{Performance evaluation of the MR-FGS and FGS algorithms on Middlebury stereo pairs without performing any post-processing. Occluded regions were not excluded for performance evaluations. MR-FGS estimates without requiring computationally expensive left-right consistency checks were more accurate than the FGS estimates.}
    \label{tab:2}       
    \begin{tabular}{cccccccc}
    \hline\noalign{\smallskip}
        Images & \multicolumn{3}{c}{FGS Final} & & \multicolumn{3}{c}{MR-FGS Final}\\
        \cline{2-4}\cline{6-8}\noalign{\smallskip} 
        &Avg.err & PSNR(dB) & Bad2.0(\%) & & Avg.err & PSNR(dB) & Bad2.0(\%)\\
        \noalign{\smallskip}\hline\noalign{\smallskip}
        Teddy      & 1.90 & 33.25 & 9.55 & & 1.69 & 33.29 & 9.24 \\
        Cones      &  2.32 & 33.33 & 15.11 & & 1.98 & 33.43 & 15.06  \\
        Dolls      &  2.04 & 32.52 & 18.98 & &  1.81 & 38.27 &16.21 \\
        Rocks1     &  2.78 & 31.10 & 12.06 & & 2.20 & 33.64 &11.91\\ 
        Motorcycle &  3.36 & 30.02 & 18.87 & & 3.09 & 30.84 & 18.25\\
    \noalign{\smallskip}\hline\noalign{\smallskip}
    \end{tabular}
\end{table}

%
\section{Conclusions}
\label{Conclusions_MRFGS}
We have presented a multi-resolution probabilistic factor-graph-based disparity estimation algorithm (MR-FGS) that improves the accuracy of disparity estimates over the FGS model using spatial dependencies as well as multi-resolution dependencies among random variables in probabilistic graphical models. We conducted extensive experiments to compare the performance of the proposed multi-resolution probabilistic factor graph model with FGS results using the Middlebury benchmark stereo datasets \cite{scharstein2003high}, \cite{scharstein2007learning}, \cite{hirschmuller2007evaluation}, \cite{scharstein2014high}. Our experimental results indicate that the multi-resolution factor graph algorithm provides disparity estimates with higher accuracy and improves contrast along the depth boundaries. In contrast to the FGS algorithm, the MR-FGS algorithm does not require the computationally expensive left-right consistency checks for the disparity estimates.

\clearpage

\bibliographystyle{IEEEtran}
\bibliography{ref.bib}

\begin{thebibliography}{10}
\providecommand{\url}[1]{#1}
\csname url@samestyle\endcsname
\providecommand{\newblock}{\relax}
\providecommand{\bibinfo}[2]{#2}
\providecommand{\BIBentrySTDinterwordspacing}{\spaceskip=0pt\relax}
\providecommand{\BIBentryALTinterwordstretchfactor}{4}
\providecommand{\BIBentryALTinterwordspacing}{\spaceskip=\fontdimen2\font plus
\BIBentryALTinterwordstretchfactor\fontdimen3\font minus
  \fontdimen4\font\relax}
\providecommand{\BIBforeignlanguage}[2]{{%
\expandafter\ifx\csname l@#1\endcsname\relax
\typeout{** WARNING: IEEEtran.bst: No hyphenation pattern has been}%
\typeout{** loaded for the language `#1'. Using the pattern for}%
\typeout{** the default language instead.}%
\else
\language=\csname l@#1\endcsname
\fi
#2}}
\providecommand{\BIBdecl}{\relax}
\BIBdecl

\bibitem{desouza2002vision}
G.~N. DeSouza and A.~C. Kak, ``Vision for mobile robot navigation: A survey,''
  \emph{IEEE transactions on pattern analysis and machine intelligence},
  vol.~24, no.~2, pp. 237--267, 2002.

\bibitem{remondino2008turning}
F.~Remondino, S.~F. El-Hakim, A.~Gruen, and L.~Zhang, ``Turning images into 3-d
  models,'' \emph{IEEE Signal Processing Magazine}, vol.~25, no.~4, pp. 55--65,
  2008.

\bibitem{shean2016automated}
D.~E. Shean, O.~Alexandrov, Z.~M. Moratto, B.~E. Smith, I.~R. Joughin,
  C.~Porter, and P.~Morin, ``An automated, open-source pipeline for mass
  production of digital elevation models (dems) from very-high-resolution
  commercial stereo satellite imagery,'' \emph{ISPRS Journal of Photogrammetry
  and Remote Sensing}, vol. 116, pp. 101--117, 2016.

\bibitem{zenati2007dense}
N.~Zenati and N.~Zerhouni, ``Dense stereo matching with application to
  augmented reality,'' in \emph{2007 IEEE International Conference on Signal
  Processing and Communications}.\hskip 1em plus 0.5em minus 0.4em\relax IEEE,
  2007, pp. 1503--1506.

\bibitem{helmer2010using}
S.~Helmer and D.~Lowe, ``Using stereo for object recognition,'' in \emph{2010
  IEEE International Conference on Robotics and Automation}.\hskip 1em plus
  0.5em minus 0.4em\relax IEEE, 2010, pp. 3121--3127.

\bibitem{scharstein2002taxonomy}
D.~Scharstein and R.~Szeliski, ``A taxonomy and evaluation of dense two-frame
  stereo correspondence algorithms,'' \emph{International journal of computer
  vision}, vol.~47, no.~1, pp. 7--42, 2002.

\bibitem{yang2008stereo}
Q.~Yang, L.~Wang, R.~Yang, H.~Stew{\'e}nius, and D.~Nist{\'e}r, ``Stereo
  matching with color-weighted correlation, hierarchical belief propagation,
  and occlusion handling,'' \emph{IEEE transactions on pattern analysis and
  machine intelligence}, vol.~31, no.~3, pp. 492--504, 2008.

\bibitem{li1994markov}
S.~Z. Li, ``Markov random field models in computer vision,'' in \emph{European
  conference on computer vision}.\hskip 1em plus 0.5em minus 0.4em\relax
  Springer, 1994, pp. 361--370.

\bibitem{ethier2009markov}
S.~N. Ethier and T.~G. Kurtz, \emph{Markov processes: characterization and
  convergence}.\hskip 1em plus 0.5em minus 0.4em\relax John Wiley \& Sons,
  2009, vol. 282.

\bibitem{shabanian2021novel}
H.~Shabanian and M.~Balasubramanian, ``A novel factor graph-based optimization
  technique for stereo correspondence estimation,'' \emph{arXiv preprint
  arXiv:2109.11077}, 2021.

\bibitem{farin2002handbook}
G.~Farin, J.~Hoschek, and M.-S. Kim, \emph{Handbook of computer aided geometric
  design}.\hskip 1em plus 0.5em minus 0.4em\relax Elsevier, 2002.

\bibitem{navarro1996image}
R.~Navarro, A.~Tabernero, and G.~Crist{\'o}bal, ``Image representation with
  gabor wavelets and its applications,'' \emph{Advances in imaging and electron
  physics}, vol.~97, pp. 2--85, 1996.

\bibitem{salem2008multiresolution}
M.~A.-M.~M. Salem, ``Multiresolution image segmentation,'' 2008.

\bibitem{shabanian2017new}
H.~Shabanian and F.~Mashhadi, ``A new approach for detecting copy-move forgery
  in digital images,'' in \emph{2017 IEEE Western New York Image and Signal
  Processing Workshop (WNYISPW)}.\hskip 1em plus 0.5em minus 0.4em\relax IEEE,
  2017, pp. 1--6.

\bibitem{tosic2005multiresolution}
I.~Tosic, I.~Bogdanova, P.~Frossard, and P.~Vandergheynst, ``Multiresolution
  motion estimation for omnidirectional images,'' in \emph{2005 13th European
  Signal Processing Conference}.\hskip 1em plus 0.5em minus 0.4em\relax IEEE,
  2005, pp. 1--4.

\bibitem{zhao2019super}
S.~Zhao, L.~Zhang, Y.~Shen, S.~Zhao, and H.~Zhang, ``Super-resolution for
  monocular depth estimation with multi-scale sub-pixel convolutions and a
  smoothness constraint,'' \emph{IEEE Access}, vol.~7, pp. 16\,323--16\,335,
  2019.

\bibitem{kim2002factorial}
J.~Kim and R.~Zabih, ``Factorial markov random fields,'' in \emph{European
  Conference on Computer Vision}.\hskip 1em plus 0.5em minus 0.4em\relax
  Springer, 2002, pp. 321--334.

\bibitem{kato2002multicue}
Z.~Kato, T.-C. Pong, and S.~G. Qiang, ``Multicue mrf image segmentation:
  Combining texture and color features,'' in \emph{Object recognition supported
  by user interaction for service robots}, vol.~1.\hskip 1em plus 0.5em minus
  0.4em\relax IEEE, 2002, pp. 660--663.

\bibitem{benedek2015multilayer}
C.~Benedek, M.~Shadaydeh, Z.~Kato, T.~Szir{\'a}nyi, and J.~Zerubia,
  ``Multilayer markov random field models for change detection in optical
  remote sensing images,'' \emph{ISPRS Journal of Photogrammetry and Remote
  Sensing}, vol. 107, pp. 22--37, 2015.

\bibitem{benedek2009detection}
C.~Benedek, T.~Szir{\'a}nyi, Z.~Kato, and J.~Zerubia, ``Detection of object
  motion regions in aerial image pairs with a multilayer markovian model,''
  \emph{IEEE Transactions on Image Processing}, vol.~18, no.~10, pp.
  2303--2315, 2009.

\bibitem{shi2007factor}
D.~Shi and J.~You, ``Factor metanetwork: a multilevel probabilistic meta-model
  based on factor graphs,'' \emph{International Journal of General Systems},
  vol.~36, no.~4, pp. 465--477, 2007.

\bibitem{zhang2019factor}
Z.~Zhang, F.~Wu, and W.~S. Lee, ``Factor graph neural network,'' \emph{arXiv
  preprint arXiv:1906.00554}, 2019.

\bibitem{gilmer2017neural}
J.~Gilmer, S.~S. Schoenholz, P.~F. Riley, O.~Vinyals, and G.~E. Dahl, ``Neural
  message passing for quantum chemistry,'' in \emph{International conference on
  machine learning}.\hskip 1em plus 0.5em minus 0.4em\relax PMLR, 2017, pp.
  1263--1272.

\bibitem{mutimbu2018factor}
L.~Mutimbu and A.~Robles-Kelly, ``A factor graph evidence combining approach to
  image defogging,'' \emph{Pattern Recognition}, vol.~82, pp. 56--67, 2018.

\bibitem{scharstein2003high}
D.~Scharstein and R.~Szeliski, ``High-accuracy stereo depth maps using
  structured light,'' in \emph{2003 IEEE Computer Society Conference on
  Computer Vision and Pattern Recognition, 2003. Proceedings.}, vol.~1.\hskip
  1em plus 0.5em minus 0.4em\relax IEEE, 2003, pp. I--I.

\bibitem{scharstein2007learning}
D.~Scharstein and C.~Pal, ``Learning conditional random fields for stereo,'' in
  \emph{2007 IEEE Conference on Computer Vision and Pattern Recognition}.\hskip
  1em plus 0.5em minus 0.4em\relax IEEE, 2007, pp. 1--8.

\bibitem{hirschmuller2007evaluation}
H.~Hirschmuller and D.~Scharstein, ``Evaluation of cost functions for stereo
  matching,'' in \emph{2007 IEEE Conference on Computer Vision and Pattern
  Recognition}.\hskip 1em plus 0.5em minus 0.4em\relax IEEE, 2007, pp. 1--8.

\bibitem{scharstein2014high}
D.~Scharstein, H.~Hirschm{\"u}ller, Y.~Kitajima, G.~Krathwohl,
  N.~Ne{\v{s}}i{\'c}, X.~Wang, and P.~Westling, ``High-resolution stereo
  datasets with subpixel-accurate ground truth,'' in \emph{German conference on
  pattern recognition}.\hskip 1em plus 0.5em minus 0.4em\relax Springer, 2014,
  pp. 31--42.

\bibitem{tomasi1998bilateral}
C.~Tomasi and R.~Manduchi, ``Bilateral filtering for gray and color images,''
  in \emph{Sixth international conference on computer vision (IEEE Cat. No.
  98CH36271)}.\hskip 1em plus 0.5em minus 0.4em\relax IEEE, 1998, pp. 839--846.

\bibitem{pearl1982reverend}
J.~Pearl, \emph{Reverend Bayes on inference engines: A distributed hierarchical
  approach}.\hskip 1em plus 0.5em minus 0.4em\relax Cognitive Systems
  Laboratory, School of Engineering and Applied Science~…, 1982.

\bibitem{kschischang2001factor}
F.~R. Kschischang, B.~J. Frey, and H.-A. Loeliger, ``Factor graphs and the
  sum-product algorithm,'' \emph{IEEE Transactions on information theory},
  vol.~47, no.~2, pp. 498--519, 2001.

\bibitem{barber2012bayesian}
D.~Barber, \emph{Bayesian reasoning and machine learning}.\hskip 1em plus 0.5em
  minus 0.4em\relax Cambridge University Press, 2012.

\bibitem{murphy2013loopy}
K.~Murphy, Y.~Weiss, and M.~I. Jordan, ``Loopy belief propagation for
  approximate inference: An empirical study,'' \emph{arXiv preprint
  arXiv:1301.6725}, 2013.

\bibitem{jain1991unsupervised}
A.~K. Jain and F.~Farrokhnia, ``Unsupervised texture segmentation using gabor
  filters,'' \emph{Pattern recognition}, vol.~24, no.~12, pp. 1167--1186, 1991.

\bibitem{shi1994good}
J.~Shi \emph{et~al.}, ``Good features to track,'' in \emph{1994 Proceedings of
  IEEE conference on computer vision and pattern recognition}.\hskip 1em plus
  0.5em minus 0.4em\relax IEEE, 1994, pp. 593--600.

\bibitem{burt1983edward}
P.~J. Burt, ``Edward, and eh adelson. the laplacian pyramid as a compact image
  code,'' \emph{IEEE Transactions on Communications}, vol.~31, no. 532-540, p.
  340, 1983.

\bibitem{brownrigg1984weighted}
D.~R. Brownrigg, ``The weighted median filter,'' \emph{Communications of the
  ACM}, vol.~27, no.~8, pp. 807--818, 1984.

\end{thebibliography}

\end{document}